\documentclass[10pt,twocolumn,letterpaper]{article}

\usepackage{iccv}
\usepackage{times}
\usepackage{epsfig}
\usepackage{graphicx}
\usepackage{amsmath}
\usepackage{amssymb}

\usepackage{kotex}
\usepackage{bm}
\usepackage{array}
\usepackage{bbm}
\usepackage{booktabs} 
\usepackage{multirow}
\usepackage{makecell}

\usepackage{colortbl}
\usepackage{xcolor}

\makeatletter
\newcommand{\ccell}[3][]{%
  \kern-\fboxsep
  \if\relax\detokenize{#1}\relax
    \expandafter\@firstoftwo
  \else
    \expandafter\@secondoftwo
  \fi
  {\colorbox{#2}}%
  {\colorbox[#1]{#2}}%
  {#3}\kern-\fboxsep
}
\makeatother
\definecolor{cellgray}{gray}{0.9}

\usepackage[pagebackref=true,breaklinks=true,letterpaper=true,colorlinks,bookmarks=false]{hyperref}

\makeatletter
\newcommand{\thickhline}{%
    \noalign {\ifnum 0=`}\fi \hrule height 1pt
    \futurelet \reserved@a \@xhline
}
\makeatother

\iccvfinalcopy 

\def\iccvPaperID{1904} 

\ificcvfinal\fi

\begin{document}

\title{Revisiting Self-Supervised Monocular Depth Estimation}

\author{Ue-Hwan Kim, Jong-Hwan Kim\\
School of Electrical Engineering, KAIST\\
Daejeon, Republic of Korea\\
{\tt\small \{uhkim, johkim\}@rit.kaist.ac.kr}
}

\maketitle
\ificcvfinal\thispagestyle{empty}\fi

\begin{abstract}
Self-supervised learning of depth map prediction and motion estimation from monocular video sequences is of vital importance---since it realizes a broad range of tasks in robotics and autonomous vehicles. A large number of research efforts have enhanced the performance by tackling illumination variation, occlusions, and dynamic objects, to name a few. However, each of those efforts targets individual goals and endures as separate works. Moreover, most of previous works have adopted the same CNN architecture, not reaping architectural benefits. Therefore, the need to investigate the inter-dependency of the previous methods and the effect of architectural factors remains. To achieve these objectives, we revisit numerous previously proposed self-supervised methods for joint learning of depth and motion, perform a comprehensive empirical study, and unveil multiple crucial insights. Furthermore, we remarkably enhance the performance as a result of our study---outperforming previous state-of-the-art performance.
   
\end{abstract}

\section{Introduction}\label{sec:introduction}
Accurate estimation of 3D structure (e.g., dense depth map) and camera motion plays a crucial role in computer vision---since it provides key grounds for a wide range of tasks in robotics, autonomous vehicles and augmented reality. Among various sensing modalities for the estimation such as LiDAR \cite{ma2019self} and stereo camera \cite{godard2017unsupervised}, monocular cameras are drawing increasing attention due to their cost-efficiency and easy-to-deploy nature. Traditional approaches identify the same points in consecutive image frames for the estimation, which demands a lot of manual feature engineering \cite{gordon2019depth}. Recently, the field is moving towards deep learning \cite{godard2019digging}. Deep neural networks learn the estimation from data without manual feature engineering and display superior performance.

In the early stage, fully-supervised methods for depth learning from single images using neural networks have emerged \cite{eigen2014depth}. However, the scarcity of datasets with ground-truth for supervision signals has triggered the development of self-supervised methods. Self-supervised learning methods jointly learn depth and motion from monocular video sequences by utilizing geometric constraints as the source of supervision signals \cite{zhou2017unsupervised}. The supervision signals come from projecting temporal frames onto nearby frames using the estimated depth and motion, reconstructing the original frames from nearby frames, and enforcing photometric consistency over the reconstructed frames.

Contemporary approaches of self-supervised monocular depth and motion learning aim to tackle a set of issues to enhance the performance: depth representation \cite{zhou2017unsupervised, godard2019digging, li2020unsupervised}, illumination variation \cite{godard2017unsupervised, yang2020d3vo}, occlusion induced by view variation \cite{godard2019digging, gordon2019depth} and dynamic objects \cite{li2019learning, li2020unsupervised}. These issues hinder the learning process as they restrict neural networks from optimally processing information or violate the assumptions implicitly imposed by the geometric constraints such as the static scene or the Lambertian surface assumptions. Although prior works have dramatically enhanced the performance compared to the early works, contemporary research on self-supervised monocular depth and motion learning possesses two main drawbacks:
\begin{itemize}
    \item Each of the research efforts stands as a separate work and the inter-dependency between them is not well known. As a result, the research on self-supervised depth and motion learning has not yet fully achieved the potential synergies between proposed methods.
    \item Conventional works, in general, have adopted the identical CNN architecture for a fair comparison of proposed learning methods. Consequently, the need to investigate the effect of architectural factors and reap architectural benefits remains.
\end{itemize}

To overcome the limitations mentioned above, we propose to have a closer look at potential synergies between various depth and motion learning methods and CNN architectures. For this, we revisit a notable subset of previously proposed learning approaches and categorize them into four classes: depth representation, illumination variation, occlusion and dynamic objects. Next, we design a comprehensive empirical study to unveil the potential synergies and architectural benefits. To cope with the large search space, we take an incremental approach in designing our experiments. As a result of our study, we uncover a number of vital insights. We summarize the most important as follows: (1) choosing the right depth representation substantially improves the performance, (2) not all learning approaches are universal, and they have their own context, (3) the combination of auto-masking and motion map handles dynamic objects in a robust manner, (4) CNN architectures influence the performance significantly, and (5) there would exist a trade-off between depth consistency and performance enhancement. Moreover, we obtain new state-of-the-art performance in the process of performing our study.


\section{Related Works}\label{sec:related_works}
In this section, we review previous research outcomes relevant to our study. We discuss the main ideas of previous works and highlight the novelty of them compared to conventional methods.

\subsection{Self-Supervised Depth Estimation}
\textbf{Monocular Depth Estimation.}
Since the estimation of depth from monocular images is an inherently ill-posed problem, early works have focused on learning the estimation with ground-truth depth labels \cite{eigen2014depth}. Nonetheless, the expensive cost of collecting ground-truth depth labels has led to the development of self-supervised monocular depth estimation \cite{zhou2017unsupervised}. In self-supervised monocular depth estimation, two CNNs learn to estimate depth and relative camera motion. Supervision signals come from warping temporal image frames onto nearby frames using the estimated depth and motion, reconstructing the original frames from nearby frames and enforcing photometric consistency. This photometric consistency implicitly assumes that scenes are static (rigid) and Lambertian. Thus, violation of the assumptions---which frequently occurs in natural scenes---degrades the depth estimation performance. Various research teams have attempted to enhance the monocular depth estimation performance in real-world scenes. Mahjourian \emph{et al.} leveraged the point cloud alignment in addition to the photometric consistency constraint \cite{mahjourian2018unsupervised}. Casser \emph{et al.} utilized semantic information to handle 3D object motion \cite{casser2019depth}. Bian \emph{et al.} pursued scale consistency to overcome the scale ambiguity problem of monocular depth and motion estimation \cite{bian2019unsupervised}. Godard \emph{et al.} reduced the performance gap between stereo and monocular self-supervised depth estimation by proposing multi-scale estimation and per-pixel minimum reprojection loss \cite{godard2019digging}. Nevertheless, these approaches still need to resolve the gradient deterioration problem caused by dynamic objects in the scenes. 

\textbf{Handling of Dynamic Objects.} One of the challenges self-supervised depth and motion estimation faces is the performance impairment due to dynamic objects. Dynamic objects hinder the evaluation of the photometric consistency constraint, corrupt supervision signals and deteriorate the performance. One straightforward approach is incorporating semantic information and handling dynamic objects. SIGNet fuses semantic information from pre-trained networks to make depth and flow predictions consistent over dynamic regions of the scene \cite{meng2019signet}. Casser \emph{et al.} and Godon \emph{et al.} utilized instance segmentation masks from pre-trained segmentation networks to handle 3D object motion \cite{casser2019depth, gordon2019depth}. Furthermore, learning optical flow as an auxiliary task would enhance the depth estimation performance. Early works introducing unsupervised learning of optical flow trained CNNs with image synthesis and local flow smoothness \cite{jason2016back, ren2017unsupervised}. Later, researchers integrated the optical flow learning into the depth estimation learning \cite{yin2018geonet, zou2018df, luo2020every}. Recently, Li \emph{et al.} proposed the group smoothness loss and the $L_{1/2}$ sparsity loss for learning dynamic object masks in a self-supervised manner \cite{li2020unsupervised}.

\textbf{Uncertainty Estimation.} Proper quantification of uncertainty prevents overconfident incorrect predictions and aids the process of decision making in real-world computer vision applications \cite{lakshminarayanan2017simple}. Uncertainties mainly arise from two sources: input data (aleatoric uncertainty) or model itself (epistemic uncertainty) \cite{kendall2017uncertainties}. The noise in data---generally originates from sensor imperfection---causes aleatoric uncertainty while the limited model capacity or unbalanced distribution of training data brings about epistemic uncertainty \cite{poggi2020uncertainty}. Researchers have actively explored this aspect even prior to the deep learning era \cite{szeliski1990bayesian, kosaka1992fast}. Currently, estimating uncertainty for neural networks is under active research. Since neural networks consist of large numbers of parameters, various methods have emerged to estimate the posterior distribution of network predictions. Examples of such methods include Bayesian neural networks \cite{ghosh2018structured, kwon2020uncertainty}, variational inference \cite{graves2011practical, mnih2014neural, blier2018description}, bootstrapped ensembles \cite{lakshminarayanan2017simple}, Monte Carlo dropout \cite{gal2016dropout, mukhoti2018evaluating} and Laplace approximation \cite{ritter2018scalable}. Recently, uncertainty estimation for self-supervised monocular depth learning has started to draw attention as well \cite{yang2020d3vo, poggi2020uncertainty, johnston2020self}.

\subsection{CNN Architectures}
\textbf{Architectures for Accuracy.} Since AlexNet demonstrated the superior performance of CNNs by winning the 2012 ImageNet competition \cite{krizhevsky2012imagenet}, a lot of research efforts have followed to improve the performance of CNNs further. To accomplish this goal, researchers have designed bigger and more accurate CNN architectures: GoogleNet \cite{szegedy2015going}, winner of the 2014 ImageNet, achieves 74.8\% top-1 accuracy with 6.8M parameters; ResNet \cite{he2016deep}, winner of the 2015 ImageNet, achieves 78.6\% top-1 accuracy with 60M parameters; SENet \cite{hu2018squeeze}, winner of the 2017 ImageNet, achieves 82.7\% top-1 accuracy with 145M parameters; recent GPipe \cite{huang2018gpipe} achieves 84.3\% top-1 accuracy with 557M parameters. Though designed for ImageNet, these architectures display satisfactory performance in other tasks and better performance in ImageNet results in better performance in other tasks \cite{he2016deep, tan2019mnasnet}. Moreover, the features learned through ImageNet get effectively-transferred to downstream tasks \cite{kornblith2019better} signifying the importance of the ImageNet research.

\textbf{Architectures for Efficiency.} After a series of breakthroughs in the ImageNet challenge over the last decade, the gigantic architectures hitting the hardware memory limit gave rise to research on CNN efficiency. For example, model compression aims to reduce model dimensions by weighing between efficiency and accuracy \cite{yang2018netadapt}. CNN architectures such as MobileNet \cite{howard2019searching} and ShuffleNet \cite{ma2018shufflenet} that can run on mobile smartphones emerged with the development of mobile devices equipped with moderate computation power. Neural architecture search automates the process of neural architecture design and reports better efficiency than hand-crafted architectures \cite{ying2019bench, wan2020fbnetv2}. Recently, EfficientNet \cite{tan2019efficientnet} has achieved both high efficiency and accuracy by closely examining scaling factors of CNNs---depth, width and resolution.

\section{Methodology}\label{sec:methodology}
In this section, we describe our study setup. First, we delineate four classes of learning approaches considered in this work. Next, we introduce CNN architectures investigated in our empirical study.

\subsection{Learning Approaches}
\textbf{Depth Representation.} Since the depth values in real-world applications are much larger than those neural networks can stably produce, appropriate depth representation enhances the performance significantly. Thus, the appropriate choice of depth representation for aiding feature representation learning plays an essential role in self-supervised monocular depth and motion learning. We compare three depth representations in our study.
\begin{itemize}
    \item \emph{Disparity} \cite{zhou2017unsupervised} is simply the inverse of depth as follows: \begin{equation}
        d = 1 / x,
    \end{equation}
    where $x$ is the output of neural networks and $d$ is the recovered depth. Due to the inversion of the depth values, it can represent distant objects stably. However, its stability suffers when adjacent objects appear in the scene ($d << 1$).
    \item \emph{Scaled disparity} \cite{godard2019digging} scales the predicted disparity to a pre-defined range to improve numerical stability. The minimum and maximum values the final depth values can take are hyper-parameters and generally set as 0.1 and 100, respectively. The scaled disparity converts the predicted disparity as follows: \begin{equation}
        x' = \sigma_\text{min} + (\sigma_\text{max} - \sigma_\text{min}) \cdot x,
    \end{equation}
    where $\sigma_\text{max}$ and $\sigma_\text{min}$ are the maximum and minimum values disparity can take. Finally, the depth values are $d = 1/x'$.
    \item \emph{Softplus} \cite{li2020unsupervised} representation lets neural networks directly predict depth values rather than disparities as follows: \begin{equation}
        d = \log (\exp(x) + 1).
    \end{equation}
    The softplus representation avoids the case where the predicted depth values equal to zeros. This setting eliminates the need for setting hyper-parameters such as minimum disparity and lets CNNs learn the optimal values from data.
\end{itemize}

\textbf{Illumination Variation.} The geometric constraints, i.e., the photometric consistency loss, implicitly assume that scenes are Lambertian. However, the illumination in real-world scenes varies according to different camera angles and different time steps. Therefore, violation of this assumption---corrupting the supervision signals---frequently occurs and needs careful handling. We consider three methods for accommodating this discrepancy.
\begin{itemize}
    \item \emph{Brightness transformation} \cite{yang2020d3vo} models the change of the image intensity between two scenes with an affine transformation as follows \begin{equation}
        \bm{I}' = a \cdot \bm{I} + b,
    \end{equation}
    where $a\! \in \! \mathbb{R}$ and $b\! \in \! \mathbb{R}$ are two parameters of the affine transformation. Despite its simplicity, this formulation effectively enhances the performance of depth and motion estimation \cite{wang2017stereo, yang2020d3vo}. Neural networks can learn the two parameters in a self-supervised manner.
    \item \emph{Structural similarity (SSIM)} \cite{godard2017unsupervised} is another remedy for the violation of the Lambertian assumption. It computes structural similarity rather than pixel similarity between two image patches as follows: \begin{equation}
        SSIM(x, y) = \frac{(2\mu_x \mu_y + c_1)(2\mu_{xy} + c_2)}{(\mu_x^2 + \mu_y^2 + c_1)(\sigma_x + \sigma_y + c_2)},
    \end{equation}
    where $x$ and $y$ are image patches, $\mu$ and $\sigma$ are patch mean and variance, respectively, $c_1=0.01^2$ and $c_2 = 0.03^2$. SSIM has become a de facto approach for self-supervised depth and motion learning; the objective function generally combines the SSIM loss and the L1 photometric loss with the ratio of $0.85:0.15$.
    \item \emph{Depth-error weighted (DW) SSIM} \cite{li2020unsupervised} penalizes the image segments where the predicted depth values are not consistent over multiple observations. The depth-error weight at a pixel position $(i, j)$ is calculated as follows: \begin{equation}
        \bm{w}^{(i,j)} = \frac{\sigma_{d}^2}{\sigma_{d}^2 + (\hat{\bm{d}}_{t, recon}^{(i,j)} - \hat{\bm{d}}_{t}^{(i,j)})^2},
    \end{equation}
    where $\hat{\bm{d}}_{t}$ is the predicted depth at time step $t$, $\hat{\bm{d}}_{t, recon}$ is the reconstructed depth map at $t$ from the predicted depth maps of nearby images and $\sigma_{d}^2 = \frac{1}{N}\sum_{i,j}(\hat{\bm{d}}_{t, recon} - \hat{\bm{d}}_{t})^2$, where $N$ is the total number of valid pixels.
\end{itemize}

\textbf{Occlusion.} Due to variations in observation angles, occlusions occur. Occlusions hinder the image reconstruction process by making parts of pixels not visible. This could induce a high photometric penalty even for the pixels with correctly-estimated depth values---corrupting the supervision signals. We examine two methods for handling occlusions.
\begin{itemize}
    \item \emph{Minimum reprojection (MR)} \cite{godard2019digging} handles occlusions by taking a minimum operation rather than an averaging operation over multiple reconstruction error maps and calculating the per-pixel photometric loss. The MR loss also handles out-of-view pixels due to ego-motion at image boundaries. The MR loss gets calculated as follows:
    \begin{equation}
        L_{mr} = \min_{t'} pe(\bm{I}_{t}, \bm{I}_{t' \rightarrow t}),
    \end{equation}
    where $pe(\bm{I}_a, \bm{I}_b)=\frac{\alpha}{2}(1-SSIM(\bm{I}_a, \bm{I}_b)) + (1-\alpha)||\bm{I}_a - \bm{I}_b||_1$.
    \item \emph{Depth consistency (DC)} \cite{gordon2019depth} utilizes the fact that the depth value at a pixel becomes multi-valued when occlusion occurs. To check the consistency, this approach projects the predicted depth $z_{ij}$ at a pixel position $(i, j)$ in the source frame, obtains the respective point $(x_{ij}, y_{ij}, z_{ij})$ in space, and applies the camera motion to reach $(i', i', z_{i'j'}')$. Then, the transformed source depth $z_{ij}'$ gets compared to the target depth $z_{i'j'}^t$ and the photometric consistency loss only counts the pixels where $z_{i'j'}' \leq z_{i'j'}^t$. This approach is not symmetric, requiring switching of source and target roles. 
\end{itemize}

\textbf{Dynamic Objects.} Dynamic objects violate the static scene assumption and gradients significantly deteriorate. Learning to handle dynamic objects in a self-supervised fashion is challenging and has become one of the major issues. We compare three methods for handling dynamic objects which do not require an external knowledge base or pre-trained networks.
\begin{itemize}
    \item \emph{Auto-masking (AM)} \cite{godard2019digging} effectively removes the scenes where the camera does not move and the objects that move with the same velocity as the camera. These entities appear as holes of infinite depth when not dealt with appropriately. Though auto-masking does not handle all types of dynamic objects, it efficiently enhances the performance by tackling the two major issues. The binary mask for auto-masking is defined as follows:
    \begin{equation}
        \mu = [\min_{t'} pe(\bm{I}_{t}, \bm{I}_{t' \rightarrow t}) < \min_{t'} pe(\bm{I}_{t}, \bm{I}_{t'})].
    \end{equation}
    \item \emph{Uncertainty modeling}---though it can account for other factors such as non-Lambertian surfaces---is one of the earliest methods for handling moving objects \cite{zhou2017unsupervised, klodt2018supervising}. Among various modeling options, contemporary approaches commonly employ the heteroscedastic aleatoric uncertainty \cite{kendall2017uncertainties}---regarding dynamic objects as observation noise---as follows:
    \begin{equation}
        L = \frac{\min_{t'} pe(\bm{I}_{t' \rightarrow t}, \bm{I}_{t})}{\sum_{t}} + \log \textstyle\sum_{t},
    \end{equation}
    where $\textstyle\sum_{t}$ is the uncertainty map of $\bm{I}_{t}$
    \item \emph{Motion map (MM)} formulation \cite{li2020unsupervised} learns a 3D object motion map $T_{\text{motion}} \in \mathbb{R}^{H \times W \times 3}$ in addition to global motion vectors consisting of translation ($\bm{t} \in \mathbb{R}^3$) and rotation ($\bm{r} \in \mathbb{R}^3$). Motion map can theoretically account for all types of dynamic objects with rigid translations in arbitrary directions. The key to learning a motion map in a self-supervised way is the $L_{1/2}$ sparsity loss, which is defined as
    \begin{equation}
        L_{1/2} = 2 \sum_{i \in \{x, y, z\}} \langle |T_i| \rangle \iint \sqrt{1+{|T_i|}/{\langle |T_i| \rangle}} dudv,
    \end{equation}
    where $\langle |T_i| \rangle$ is the spatial average of $T_i$.
    In addition, the motion map approach should be applied after a few training epochs and fed with estimated depth maps for stable learning.
\end{itemize}

\subsection{CNN Architectures}
\begin{table}[!t]
\centering
\renewcommand{\arraystretch}{1.2}
\caption{CNN architectures employed in our study. We investigate three types of CNN architectures and ten variants in total. The numbers in the table are from our implementation and the pre-trained weights we utilize.}
\label{tb:cnn_architectures}
\vskip 0.05in
\begin{small}
\begin{tabular}{l c c c}
\thickhline
\multicolumn{1}{c}{\textbf{Architecture}} & \textbf{\#Parameters} & \textbf{\#FLOPs} & \textbf{\makecell{top-1\\ acc (\%)}}\\
\toprule
ResNet-18 & 11.4M & 1.8B & 69.76\\
ResNet-50 & 23.9M & 4.1B & 76.13\\
ResNet-101 & 42.8M & 7.9B & 77.37\\
\midrule
DeResNet-18 & 11.4M & 1.9B & -\\
DeResNet-50 & 23.9M & 4.3B & 78.2\\
DeResNet-101 & 42.8M & 8.2B & 79.2\\
\midrule
EfficientNet-B0 & 5.3M & 0.4B & 76.3\\
EfficientNet-B1 & 7.8M & 0.7B & 78.8\\
EfficientNet-B2 & 9.2M & 1.0B & 79.8\\
EfficientNet-B4 & 19.3M & 4.2B & 82.6\\
\bottomrule
\end{tabular}
\end{small}
\end{table}

Table \ref{tb:cnn_architectures} summarizes the CNN architectures utilized in our study. We investigate three types of CNN architectures and different scales of them. We select the architectures extensively-verified in the literature and the scales considering both the number of parameters and computational complexities. In total, we examine ten variant CNN architectures with different scales and types.

\textbf{ResNet} is one of the most broadly-applied CNN architectures due to its effectiveness and efficiency over various application areas \cite{he2016deep}. Its main blocks consist of a $7 \times 7$ convolution layer, a series of four residual blocks, and a pooling layer followed by a fully-connected layer. Each residual block contains different number of residual units in the form of $y:=x + \mathcal{F}(x)$, where $\mathcal{F}$ represents a residual function composed of a set of convolutions, ReLU non-linearities \cite{nair2010rectified} and batch normalization layers \cite{ioffe2015batch}. Adjusting the number of layers in each residual block offers a way to scale-up or scale-down the network depth. We utilize three scales of ResNet, namely ResNet-18, ResNet-50 and ResNet-101, and investigate the architectural effect on the depth learning performance. Furthermore, we cut out the components of ResNet after the series of four residual blocks for our experimental purpose.

\textbf{Deformable ResNet (DeResNet)} replaces ordinary convolution layers in ResNet with deformable convolution layers \cite{dai2017deformable, zhu2019deformable}. Deformable convolution layers allow free-form sampling rather than ordinary grid sampling, which leads to performance enhancement in several tasks such as object detection, semantic segmentation and instance segmentation. This flexible sampling scheme would foster CNNs to learn better features for depth and motion estimation as well. We follow the design choice introduced in \cite{zhu2019deformable} and replace each of the $3 \times 3$ convolution layers in ResNet with a $3 \times 3$ deformable convolution layer. However, we do not adopt the modulated deformable modules to minimize the computational time overhead. We utilize three scales of DeResNet: DeResNet-18, DeResNet-50, and DeResNet-101. As in the case of ResNet, we just utilize the feature extraction part of DeResNet.

\textbf{EfficientNet} has achieved both efficiency and accuracy across various application areas \cite{tan2019efficientnet}. The key to designing the EfficientNet architecture is a compound scaling method. The scaling method systematically scales all dimensions of CNN depth, width and resolution with a set of fixed scaling coefficients. Moreover, the design of the baseline network, EfficientNet-B0, was originated from a neural architecture search using the AutoML Mnas framework \cite{tan2019mnasnet}. The resulting network consists of mobile inverted bottleneck residual convolution (MBConv) blocks similar to MobileNetV2 \cite{sandler2018mobilenetv2} and MnasNet \cite{tan2019mnasnet} in addition to optional squeeze-and-excitation blocks \cite{hu2018squeeze}. We utilize four scales of EfficientNet: EfficientNet-B0, EfficientNet-B1, EfficientNet-B2, and EfficientNet-B4.

\section{Experiments}\label{sec:experiments}
In this section, we delineate the experiment settings and implementation details for our study. Then, we present the experiment results\footnote{We present qualitative results and the comparison of the proposed method to the latest models in the supplementary material.} and analyze them to disclose crucial insights for self-supervised monocular depth learning.

\subsection{Settings}
\textbf{Dataset.} We conduct our study using the KITTI2015 dataset \cite{geiger2013vision}. The dataset comprises car driving scenes in outdoor environments. A number of dynamic objects appearing in the scenes and fast car movement make the dataset challenging for self-supervised monocular depth learning. We follow the standard split referred to as Eigen split \cite{eigen2015predicting}. Eigen split consists of 39,810 monocular image sequences for training and 4,424 sequences for validation. During training, we remove the static scenes from the dataset. For testing the depth estimation performance, we use 697 images with ground-truth depth labels. 

\textbf{Metrics.} By following the standard evaluation protocol \cite{godard2017unsupervised}, we use the following seven standard metrics to quantitatively measure the depth estimation performance: Absolute Relative Difference (ARD) \cite{saxena2008make3d}, Squared Relative Difference (SRD), Root Mean Square Error (RMSE) \cite{li2010towards}, RMSE log \cite{eigen2014depth} and three classes of Thresholds ($\delta < \nu$, $\nu \in \{1.25, 1.25^2, 1.25^3\}$) \cite{ladicky2014pulling}.


\textbf{Ablation Study.} To manage the large search space, we take an incremental approach: we group the four categories of learning approaches into two; investigate the inter-dependency within each group; explore the architectural impacts with the leading learning configuration. First, we examine the inter-dependency between depth representation and the methods for handling illumination variations. We name the models in the first stage with the prefixes of R (reciprocal; disparity), S (scaled disparity), and L (log; softplus). Next, we investigate the relations between methods for occlusions and those for dynamic objects. Models in this stage are named with the prefixes of M (min. reprojection), D (depth consistency), and C (combined). Finally, we study architectural impact.

\subsection{Implementation Details}
\textbf{Development Environment.} We use the PyTorch library to implement the models for our experiments. We employ a single NVIDIA 2080Ti GPU for all experiments. We jointly learn the depth and pose networks through the Adam optimizer \cite{kingma2014adam} with $\beta_1=0.9$, $\beta_2=0.999$ and the learning rate of $10^{-4}$. We apply an image data augmentation composed of horizontal flips, random contrast, saturation, hue jitter and brightness.

\textbf{Data Processing.} Though the dataset contains a set of different-sized images, we use the same intrinsic matrix for all training and validation images: we place the principal point of the camera at the center of images and average all the focal lengths in the dataset to obtain a single focal length. For evaluation of the performance metrics, we cap depth to 80m following the standard protocol \cite{godard2017unsupervised}. In addition, we apply the per-image median ground-truth scaling \cite{zhou2017unsupervised} to cope with the scale ambiguity of monocular depth estimation.

\textbf{Technical Issues.} For the disparity and softplus representations, we apply the inverse of the variance of depth maps as an additional loss to stabilize the learning process \cite{li2020unsupervised}. The two representations make the networks diverge in all the cases without the variance loss. We scale the variance loss by $10^{-6}$. Moreover, we employ just the $L_{1/2}$ loss and the smoothness loss for the motion map formulation but not the cyclic consistency loss \cite{li2020unsupervised}. The cyclic consistency loss could corrupt the gradients since the motion map to compensate dynamic objects is not symmetric. Further, we select one of the feature maps of the same size from the later stages in the case of EfficientNet since the network architecture generates multiple same-sized feature maps. For instance, we sample the feature maps at the layer depth of (2, 6, 10, 22, 30) in the case of EfficientNet-B4. Next, we apply auto-masking whenever possible except for the models with the brightness transformation or/and the uncertainty modeling. These two methods employing weighting of the photometric loss cause networks to learn a unary mask rather than a binary mask---excluding all pixels. Lastly, we set the pose network as ResNet18 for all experiments.

\subsection{Learning Approaches}
\begin{table*}[t]
\caption{Inter-dependency between various learning approaches. The double-edged line in the middle separates the first and second stages. The overall leading performance metrics are in bold while the leading performance metrics in each group are underlined. Repr, DW, Occ, AM, Uncrt and MM in the table stand for representation, depth-error weighted SSIM, occlusions, auto-masking, uncertainty modeling and motion map, respectively. ${\dagger}$-indicates the previous state-of-the-art configuration \cite{godard2019digging}}
\label{tb:result_ablation_learn}
\vskip 0.05in
\begin{center}
\begin{footnotesize}
\begin{tabular}{lcccccccccccccc}
\toprule
\multicolumn{1}{l}{\multirow{2}[2]{*}{\textbf{ID}}} & \multicolumn{1}{c}{\multirow{2}[2]{*}{\textbf{Repr}}} & \multicolumn{2}{c}{\textbf{Illumination}} &  \multicolumn{1}{c}{\multirow{2}[2]{*}{\textbf{Occ}}} & \multicolumn{3}{c}{\textbf{Dynamic Object}} & \multicolumn{1}{c}{\multirow{2}[2]{*}{{ARD}}} & \multicolumn{1}{c}{\multirow{2}[2]{*}{{SRD}}} & \multicolumn{1}{c}{\multirow{2}[2]{*}{{RMSE}}} & \multicolumn{1}{c}{\multirow{2}[2]{*}{{\makecell{RMSE\\log}}}} & \multicolumn{1}{c}{\multirow{2}[2]{*}{{$\delta\!\!<\!\!1.25$}}}
& \multicolumn{1}{c}{\multirow{2}[2]{*}{{$\delta\!\!<\!\!1.25^2$}}}
& \multicolumn{1}{c}{\multirow{2}[2]{*}{{$\delta\!\!<\!\!1.25^3$}}}\\
\cmidrule(lr){3-4}
\cmidrule(lr){6-8}
& & $a\bm{I}+b$& DW  & & AM & Uncrt & MM &  &  &  &  & \\
\midrule 
R0&$1/x$ & - & - & - & - & - & - & 0.123  &   1.188  &   5.148  &   0.202  &   0.867  &   0.954  &   0.978\\
R1&$1/x$ & - & - & MR & - & - & - & 0.121  &   1.044  &   5.074  &   0.198  &   0.869  &   0.957  &   0.980\\
R2&$1/x$ & - & - & MR & \checkmark & - & - &   0.119  &   0.896  &   4.882  &   0.196  &   0.870  &   \underline{0.958}  &   \underline{0.981}\\
R3&$1/x$ & \checkmark & - & MR & - & - & - &   0.122  &   1.050  &   5.083  &   0.199  &   0.869  &   0.957  &   0.980\\
R4&$1/x$ & - & \checkmark & MR & \checkmark & - & - &   \underline{0.118}  &   \underline{0.872}  &  \underline{ 4.805}  &   \underline{0.196}  &   \underline{0.872}  &   \underline{0.958}  &   0.980\\
R5&$1/x$ & \checkmark & \checkmark & MR & - & - & - &   0.120  &   1.012  &   5.027  &   0.197  &   \underline{0.872}  &   \underline{0.958}  &   0.980\\
\midrule 
S0& $(0.1,\!100)$ & - & - & - & - & - & - & 0.122  &   1.095  &   5.124  &   0.202  &   0.868  &   0.954  &   0.978\\
S1& $(0.1,\!100)$ & - & - & MR & - & - & - &   0.121  &   1.052  &   5.071  &   0.198  &   0.871  &   0.957  &   \underline{0.980}\\
S2$^{\dagger}$& $(0.1,\!100)$ & - & - & MR & \checkmark & - & - & \underline{0.117}  &   \underline{0.899}  &   \underline{4.882}  &   \underline{0.196}  &   0.872  &   \underline{0.958}  &  \underline{0.980}\\
S3& $(0.1,\!100)$ & \checkmark & - & MR & - & - & - &   0.121  &   1.014  &   5.044  &   0.198  &   0.867  &   0.957  &   \underline{0.980}\\
S4& $(0.1,\!100)$ & - & \checkmark & MR & \checkmark & - & - &   0.121  &   0.938  &   4.933  &   0.199  &   0.868  &   0.956  &   \underline{0.980}\\
S5& $(0.1,\!100)$ & \checkmark & \checkmark & MR & - & - & - &   0.118  &   1.002  &   5.055  &   0.197  &   \underline{0.873}  &   0.957  &   \underline{0.980}\\
\midrule 
L0&$\log$  & - & - & - & - & - & - & 0.131 & 1.446 & 5.403 & 0.211 & \underline{0.963} & 0.951 & 0.976\\
L1&$\log$  & - & - & MR & - & - & - & 0.120  &   0.959  &   4.971  &   0.197  &   0.871  &   \underline{0.958}  &   0.980\\
L2&$\log$  & - & - & MR & \checkmark & - & - &  \underline{0.116} &   \underline{0.866}  &   \underline{4.884}  &   \underline{0.196}  &   0.874  &   \underline{0.958}  &   \underline{0.981}\\
L3&$\log$  & \checkmark & - & MR & - & - & - &   0.120  &   1.027  &   5.040  &   0.197  &   0.871  &   0.957  &   0.980\\
L4&$\log$  & - & \checkmark & MR & \checkmark & - & - &   0.121  &   0.954  &   4.953  &   0.199  &   0.866  &   0.957  &   0.979\\
L5&$\log$  & \checkmark & \checkmark & MR & - & - & - &   0.120  &   1.001  &   5.034  &   0.197  &   0.872  &   0.957  &   0.980\\
\midrule\midrule 
M0&$\log$  & - & - & MR & - & - & - & 0.120  &   0.959  &   4.971  &   0.197  &   0.871  &   0.958  &   0.980\\
M1&$\log$  & - & - & MR & \checkmark & - & - &  0.116 &   0.866  &   4.884  &   0.196  &   0.874  &   0.958  &   0.981\\
M2&$\log$ & - & - & MR & - & \checkmark & - &   0.125  &   0.940  &   5.010  &   0.198  &   0.853  &   0.954  &   0.982 \\
\cellcolor[gray]{0.9}M3&\cellcolor[gray]{0.9}$\log$ & \cellcolor[gray]{0.9}- & \cellcolor[gray]{0.9}- & \cellcolor[gray]{0.9}MR & \cellcolor[gray]{0.9}\checkmark & \cellcolor[gray]{0.9}- & \cellcolor[gray]{0.9}\checkmark & \cellcolor[gray]{0.9}\textbf{\underline{0.114}}  &   \cellcolor[gray]{0.9}\textbf{\underline{0.825}}  &   \cellcolor[gray]{0.9}\textbf{\underline{4.706}}  &   \cellcolor[gray]{0.9}\textbf{\underline{0.191}}  &   \cellcolor[gray]{0.9}\textbf{\underline{0.877}}  &   \cellcolor[gray]{0.9}\textbf{\underline{0.960}}  &   \cellcolor[gray]{0.9}0.982\\
M4&$\log$  & - & - & MR & - & \checkmark & \checkmark & 0.126  &   0.857  &   4.954  &   0.197  &   0.843  &   0.953  &   \textbf{\underline{0.983}}\\
\midrule 
D0&$\log$  & - & - & DC & - & - & - &   0.124  &   1.046  &   5.069  &   0.203  &   0.864  &   0.953  &   0.978\\
D1&$\log$  & - & - & DC & \checkmark & - & - & 0.119  &   0.919  &   4.925  &   \underline{0.198}  &   \underline{0.867}  &   \underline{0.955}  &   0.980\\
D2&$\log$  & - & - & DC & - & \checkmark & - & 0.129  &   0.924  &   5.068  &   0.203  &   0.840  &   0.949  &   \underline{0.981}\\
D3&$\log$  & - & - & DC & \checkmark & - & \checkmark & \underline{0.117}  &   \underline{0.845}  &   \underline{4.918}  &   0.199  &   0.865  &   0.954  &   0.980\\
D4&$\log$  & - & - & DC & - & \checkmark & \checkmark & 0.134  &   0.971  &   5.870  &   0.216  &   0.811  &   0.936  &   0.978\\
\midrule 
C0&$\log$  & - & - & M+D & - & - & - &   0.118  &   0.964  &   4.960  &   \underline{0.195}  &   \underline{0.871}  &   \underline{0.958}  &   \underline{0.981}\\
C1&$\log$  & - & - & M+D & \checkmark & - & - & 0.118  &   0.878  &   \underline{4.842}  &   0.196  &   0.867  &   0.957  &   \underline{0.981}\\
C2&$\log$  & - & - & M+D & - & \checkmark & - & 0.127  &   0.993  &   5.024  &   0.198  &   0.854  &   0.954  &   \underline{0.981}\\
C3&$\log$  & - & - & M+D & \checkmark & - & \checkmark & \underline{0.116}  &   \underline{0.878}  &   4.931  &   \underline{0.195}  &   0.869  &   0.957  &   \underline{0.981} \\
C4&$\log$  & - & - & M+D & - & \checkmark & \checkmark & 0.136  &   1.069  &   6.332  &   0.224  &   0.811  &   0.931  &   0.974\\
\bottomrule
\end{tabular}
\end{footnotesize}
\end{center}
\end{table*}

\begin{table*}[t]
\caption{The effect of CNN architectures on the monocular depth estimation performance. A pertinent configuration of the architecture enhances the performance (DeResNet-50). All models display satisfactory processing speeds.}
\label{tb:result_ablation_arch_depth}
\vskip 0.05in
\begin{center}
\begin{small}
\begin{tabular}{llcccccccc}
\toprule
\multicolumn{1}{c}{\textbf{Architecture}} & \multicolumn{1}{c}{\textbf{Epochs}} & \textbf{ARD} & \textbf{SRD} & \textbf{RMSE} & \textbf{RMSE log} & $\delta<1.25$
& $\delta<1.25^2$
& $\delta<1.25^3$ & \textbf{FPS} \\
\midrule
ResNet-18 & $<20$ &   0.114  &   0.825  &   4.706  &   0.191  &   0.877  &   0.960  &   0.982 & \underline{\textbf{127.07}}\\
ResNet-50 & $<20$ &   \underline{0.110}  &   \underline{\textbf{0.735}}  &   \underline{4.606}  &   \underline{\textbf{0.187}}  &   \underline{0.880}  &   \underline{\textbf{0.961}}  &   \underline{\textbf{0.983}} & 62.94\\
ResNet-101 & $<20$ &   0.112  &   0.756  &   4.655  &   0.191  &   0.875  &   0.960  &   0.982 & 46.86\\
\midrule
DeResNet-18 & $<20$ &   0.130  &   0.907  &   5.014  &   0.208  &   0.845  &   0.948  &   0.978 & \underline{103.44}\\
\cellcolor[gray]{0.9}DeResNet-50 & \cellcolor[gray]{0.9}$<20$ &   \cellcolor[gray]{0.9}\underline{\textbf{0.108}}  &   \cellcolor[gray]{0.9}\underline{0.737}  &   \cellcolor[gray]{0.9}\underline{\textbf{4.562}}  &   \cellcolor[gray]{0.9}\underline{\textbf{0.187}}  &   \cellcolor[gray]{0.9}\underline{\textbf{0.883}}  &   \cellcolor[gray]{0.9}\underline{\textbf{0.961}}  &   \cellcolor[gray]{0.9}\underline{0.982} & \cellcolor[gray]{0.9}68.92\\
DeResNet-101 & $<20$ &   0.114  &   0.832  &   4.752  &   0.195  &   0.876  &   0.957  &   0.980  & 54.04\\
\midrule
EfficientNet-B0 & $<5$& 0.120  &   \underline{0.804}  &   5.025  &   0.195  &   0.852  &   0.956  &   0.983 & \underline{52.61}\\
EfficientNet-B1 & $<5$& 0.140  &   0.948  &   5.541  &   0.213  &   0.811  &   0.943  &   0.981 & 51.37\\
EfficientNet-B2& $<5$&   0.124  &   0.860  &   \underline{4.732}  &   0.190  &   0.859  &   \underline{0.960}  &   \underline{0.984} & 42.48\\
EfficientNet-B4 & $<10$ &   \underline{0.113}  &   0.864  &   4.785  &   \underline{0.189}  &   \underline{0.875}  &   \underline{0.960}  &   0.982 & 37.79\\
\bottomrule
\end{tabular}
\end{small}
\end{center}
\end{table*}

\begin{figure*}[t]
\begin{center}
   \includegraphics[width=1\linewidth]{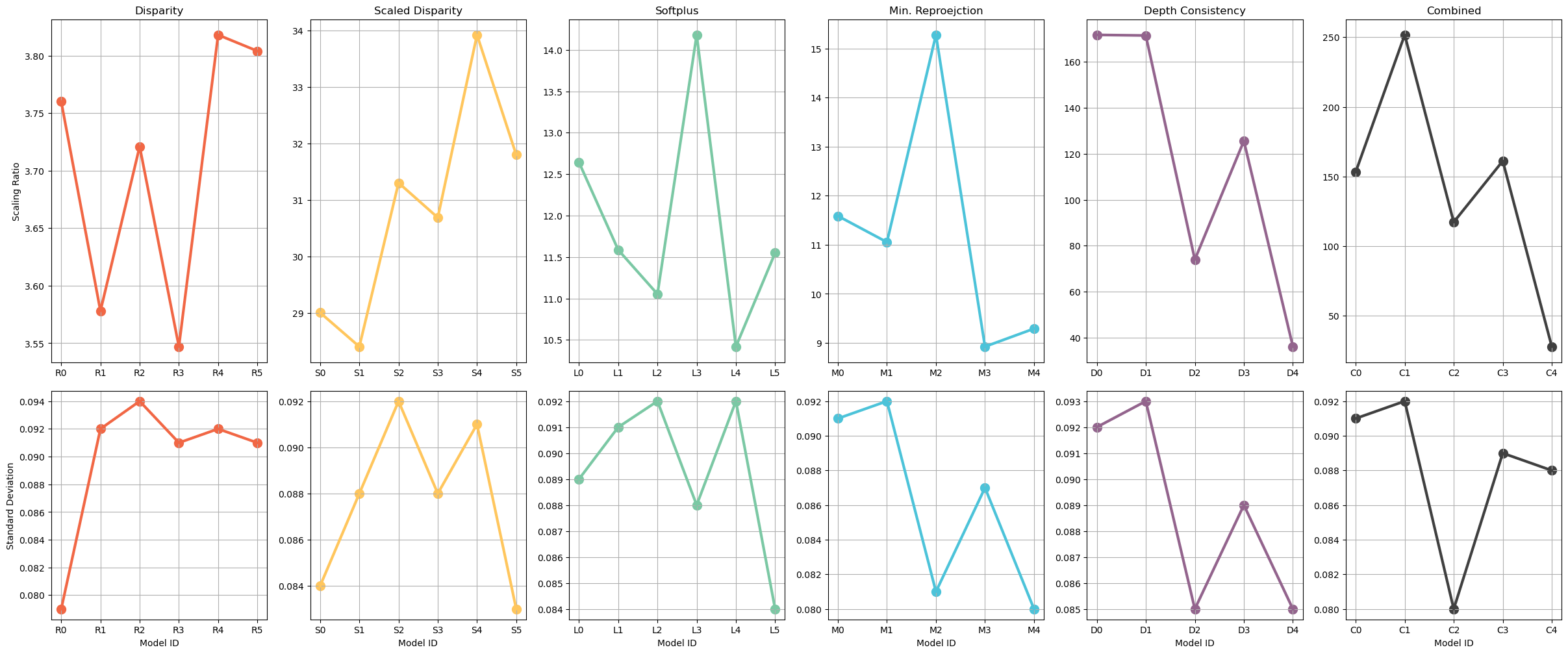}
\end{center}
   \caption{The scaling ratio and standard deviation of depth scales for each model. A large scaling ratio indicates small depth values learned, while a large standard deviation implies inconsistent-scale depth estimation. Compared to the ground-truth depth values, the models combining the MR loss and depth consistency learned much smaller depth values.}
\label{fig:scale_consistency}
\end{figure*}


\textbf{Better Depth Representation Exists.} Table \ref{tb:result_ablation_learn} summarizes the empirical study results. Among three depth representations considered in this study, the softplus representation displays superior performance. From R and S to L, the performance likely improves. This indicates that simple replacement of depth representation aids networks in learning better monocular depth estimation. Furthermore, the direct estimation of depth rather than the conventional disparity-based depth estimation shows improved performance. We presume that the softplus representation requiring zero hyper-parameters lets networks freely learn the optimal depth values. On the other hand, depth learning with the softplus representation diverged several times during our investigation, while the scaled disparity hardly showed such instability. This implies that there might exist trade-offs between representation power and learning stability at the current state.

\textbf{Not All Approaches Are Universal.} The effectiveness of several learning approaches depends on other learning approaches employed together. For instance, DW-SSIM is most effective with the disparity representation, while it does not evidently boost the performance for other representations (compare X2 and X4, where X$\in$\{R, S, L\}). Further, the brightness transformation enhances the performance with the scaled disparity to a certain degree though it is not apparent for other representations (compare X1 and X3, where X$\in$\{R, S, L\}). Similarly, the MR loss and DC adequately handle occlusions (compare L0 against M0 and D0) and the combination of the two methods seems to result in a synergy (compare M0 and D0 against C0), but the combination of the two methods does not display such synergy when associated with other learning approaches.

\textbf{Auto-Mask and Motion Map Whenever Possible.} These two approaches consistently enhance the monocular depth estimation performance (compare X0 against X1 and X3, where X$\in$\{M, D, C\}). We suppose these two methods are complementary in dealing with dynamic objects since they target exclusive sets of dynamic objects: auto-masking copes with static scenes and objects moving with the same velocity as the camera, and motion map deals with objects with substantial movements. At this stage, we mark the M3 model as our baseline model for investigating the architectural impacts. One thing to note here is that we already remarkably enhanced the performance of the previous state-of-the-art model (S2)\footnote{Though the performance of the previous state-of-the-art model is slightly lower than the performance reported in the literature \cite{godard2019digging} in our experiment, the performance of our M3 model still surpasses that of the reported performance.} by properly shaping the learning configuration.

\subsection{Architectural Impacts}
Table \ref{tb:result_ablation_arch_depth} displays the effect of CNN architectures on monocular depth estimation performance. In the case of ResNet and DeResNet, the performance improves in the order of 18, 101 and 50 layers. This implies that merely scaling the network size does not result in performance enhancement. Moreover, the experiment results show that deformable convolutions definitely learn better features for depth estimation only when adequately placed (DeResNet-50): DeResNet-101 displays inferior performance than ResNet-101, while the performance degradation in DeResNet-18 can be attributed to unavailable pre-trained weights. The DeResNet-50 model once again boosts the depth estimation performance.

Next, EfficientNet-B0, B1 and B2 diverged within five epochs of training. We attempted to train the models multiple times with different learning parameters but only to fail in all cases. On the other hand, the learning process was stable for EfficientNet-B4---although its performance saturated within ten epochs of training. We assume that the EfficientNet architecture demands another training setting for stable depth learning. Taking the small numbers of training epochs into account, the performance of EfficientNets is moderate. In summary, the architectural factors require careful incorporation with other components similar to the learning approaches whose effectiveness depends on the learning configuration. We expect the application of recent advances in neural architecture search would even further enhance the performance. 

\subsection{Scale Consistency Analysis}
Fig. \ref{fig:scale_consistency} delineates the scaling ratios (SR) and standard deviations of SR of each model. Larger scaling ratios indicate smaller depth values learned compared to the ground-truth depth values. The scaled disparity representation makes networks learn smaller depth values (SR$\approx$31). This mechanism would stabilize the learning process. The disparity (SR$\approx$3.6) and softplus (SR$\approx$12) representations learn larger depth values than the scaled disparity representation. Furthermore, the MR loss does not dramatically vary the depth value range networks learn, whereas depth consistency makes networks learn much smaller depth values (compare the 3rd against 4th and 5th graphs). We would further investigate the relationship between the scales of depth values learned and the learning stability.

Besides, the standard deviation graphs share a similar structure. First, the two local maxima (X2 and X4, where X$\in$\{R, S, L\}) in the disparity, scaled disparity and softplus graphs signal that the MR loss and AM increase the standard deviation while improving the depth estimation performance. Next, the uncertainty modeling shrinks the standard deviation while it does not significantly contribute to the depth learning (X2 and X4, where X$\in$\{M, D, C\}). This trend could imply the trade-offs between performance and depth scale consistency.
\section{Conclusion}\label{sec:discussion_conclusion}
In this work, we examined the self-supervised monocular depth and motion learning from previously unexplored angles. By designing a thorough empirical study, we revealed multiple essential insights: (1) better depth representation exists; thus, an appropriate choice of depth representation dramatically improves the performance, (2) not all learning approaches are universal, and they have their own effective setting, (3) the combination of auto-masking and motion map formulation consistently enhances the performance by handling dynamic objects, (4) CNN architectures do affect the performance substantially, and neural architecture search for self-supervised depth learning lingers, and (5) there might exist trade-offs between depth scale consistency and performance enhancement. Moreover, we obtained new state-of-the-art performance as a result of our study. Finally, we expect our extensive investigation has revealed crucial insights for future research directions for the corresponding research community.

{\small
\bibliographystyle{ieee_fullname}
\bibliography{reference}

\begin{thebibliography}{10}\itemsep=-1pt

\bibitem{bian2019unsupervised}
Jiawang Bian, Zhichao Li, Naiyan Wang, Huangying Zhan, Chunhua Shen, Ming-Ming
  Cheng, and Ian Reid.
\newblock Unsupervised scale-consistent depth and ego-motion learning from
  monocular video.
\newblock In {\em Advances in Neural Information Processing Systems (NeurIPS)},
  volume~32, pages 35--45, 2019.

\bibitem{blier2018description}
L{\'e}onard Blier and Yann Ollivier.
\newblock The description length of deep learning models.
\newblock In {\em Advances in Neural Information Processing Systems (NeurIPS)},
  pages 2220--2230, 2018.

\bibitem{casser2019depth}
Vincent Casser, Soeren Pirk, Reza Mahjourian, and Anelia Angelova.
\newblock Depth prediction without the sensors: Leveraging structure for
  unsupervised learning from monocular videos.
\newblock {\em Proceedings of the AAAI Conference on Artificial Intelligence},
  33(01):8001--8008, Jul. 2019.

\bibitem{dai2017deformable}
Jifeng Dai, Haozhi Qi, Yuwen Xiong, Yi Li, Guodong Zhang, Han Hu, and Yichen
  Wei.
\newblock Deformable convolutional networks.
\newblock In {\em Proceedings of the IEEE International Conference on Computer
  Vision (ICCV)}, pages 764--773, 2017.

\bibitem{eigen2015predicting}
David Eigen and Rob Fergus.
\newblock Predicting depth, surface normals and semantic labels with a common
  multi-scale convolutional architecture.
\newblock In {\em Proceedings of the IEEE International Conference on Computer
  Vision (ICCV)}, pages 2650--2658, 2015.

\bibitem{eigen2014depth}
David Eigen, Christian Puhrsch, and Rob Fergus.
\newblock Depth map prediction from a single image using a multi-scale deep
  network.
\newblock {\em Advances in Neural Information Processing Systems (NeurIPS)},
  27:2366--2374, 2014.

\bibitem{gal2016dropout}
Yarin Gal and Zoubin Ghahramani.
\newblock Dropout as a bayesian approximation: Representing model uncertainty
  in deep learning.
\newblock In {\em International Conference on Machine Learning (ICML)}, pages
  1050--1059. PMLR, 2016.

\bibitem{geiger2013vision}
Andreas Geiger, Philip Lenz, Christoph Stiller, and Raquel Urtasun.
\newblock Vision meets robotics: The kitti dataset.
\newblock {\em The International Journal of Robotics Research},
  32(11):1231--1237, 2013.

\bibitem{ghosh2018structured}
Soumya Ghosh, Jiayu Yao, and Finale Doshi-Velez.
\newblock Structured variational learning of bayesian neural networks with
  horseshoe priors.
\newblock In {\em International Conference on Machine Learning (ICML)}, pages
  1744--1753. PMLR, 2018.

\bibitem{godard2017unsupervised}
Cl{\'e}ment Godard, Oisin Mac~Aodha, and Gabriel~J Brostow.
\newblock Unsupervised monocular depth estimation with left-right consistency.
\newblock In {\em Proceedings of the IEEE Conference on Computer Vision and
  Pattern Recognition (CVPR)}, pages 270--279, 2017.

\bibitem{godard2019digging}
Cl{\'e}ment Godard, Oisin Mac~Aodha, Michael Firman, and Gabriel~J Brostow.
\newblock Digging into self-supervised monocular depth estimation.
\newblock In {\em Proceedings of the IEEE/CVF International Conference on
  Computer Vision (ICCV)}, pages 3828--3838, 2019.

\bibitem{gordon2019depth}
Ariel Gordon, Hanhan Li, Rico Jonschkowski, and Anelia Angelova.
\newblock Depth from videos in the wild: Unsupervised monocular depth learning
  from unknown cameras.
\newblock In {\em Proceedings of the IEEE/CVF International Conference on
  Computer Vision (ICCV)}, pages 8977--8986, 2019.

\bibitem{graves2011practical}
Alex Graves.
\newblock Practical variational inference for neural networks.
\newblock In {\em Advances in Neural Information Processing Systems (NeurIPS)},
  pages 2348--2356. Citeseer, 2011.

\bibitem{he2016deep}
Kaiming He, Xiangyu Zhang, Shaoqing Ren, and Jian Sun.
\newblock Deep residual learning for image recognition.
\newblock In {\em Proceedings of the IEEE Conference on Computer Vision and
  Pattern Recognition (CVPR)}, pages 770--778, 2016.

\bibitem{howard2019searching}
Andrew Howard, Mark Sandler, Grace Chu, Liang-Chieh Chen, Bo Chen, Mingxing
  Tan, Weijun Wang, Yukun Zhu, Ruoming Pang, Vijay Vasudevan, et~al.
\newblock Searching for mobilenetv3.
\newblock In {\em Proceedings of the IEEE/CVF International Conference on
  Computer Vision (ICCV)}, pages 1314--1324, 2019.

\bibitem{hu2018squeeze}
Jie Hu, Li Shen, and Gang Sun.
\newblock Squeeze-and-excitation networks.
\newblock In {\em Proceedings of the IEEE Conference on Computer Vision and
  Pattern Recognition (CVPR)}, pages 7132--7141, 2018.

\bibitem{huang2018gpipe}
Yanping Huang, Youlong Cheng, Ankur Bapna, Orhan Firat, Mia~Xu Chen, Dehao
  Chen, HyoukJoong Lee, Jiquan Ngiam, Quoc~V Le, Yonghui Wu, et~al.
\newblock Gpipe: Efficient training of giant neural networks using pipeline
  parallelism.
\newblock {\em arXiv preprint arXiv:1811.06965}, 2018.

\bibitem{ioffe2015batch}
Sergey Ioffe and Christian Szegedy.
\newblock Batch normalization: Accelerating deep network training by reducing
  internal covariate shift.
\newblock In {\em International Conference on Machine Learning (ICML)}, pages
  448--456. PMLR, 2015.

\bibitem{jason2016back}
J~Yu Jason, Adam~W Harley, and Konstantinos~G Derpanis.
\newblock Back to basics: Unsupervised learning of optical flow via brightness
  constancy and motion smoothness.
\newblock In {\em Proceedings of the European Conference on Computer Vision
  (ECCV)}, pages 3--10. Springer, 2016.

\bibitem{johnston2020self}
Adrian Johnston and Gustavo Carneiro.
\newblock Self-supervised monocular trained depth estimation using
  self-attention and discrete disparity volume.
\newblock In {\em Proceedings of the IEEE/CVF Conference on Computer Vision and
  Pattern Recognition (CVPR)}, pages 4756--4765, 2020.

\bibitem{kendall2017uncertainties}
Alex Kendall and Yarin Gal.
\newblock What uncertainties do we need in bayesian deep learning for computer
  vision?
\newblock In {\em Advances in Neural Information Processing Systems (NeurIPS)},
  page 5580–5590, Red Hook, NY, USA, 2017. Curran Associates Inc.

\bibitem{kim2020simvodis}
Ue-Hwan Kim, Seho Kim, and Jong-Hwan Kim.
\newblock Simvodis: Simultaneous visual odometry, object detection, and
  instance segmentation.
\newblock {\em IEEE Transactions on Pattern Analysis and Machine Intelligence},
  2020.

\bibitem{kingma2014adam}
Diederik~P Kingma and Jimmy Ba.
\newblock Adam: A method for stochastic optimization.
\newblock {\em arXiv preprint arXiv:1412.6980}, 2014.

\bibitem{klodt2018supervising}
Maria Klodt and Andrea Vedaldi.
\newblock Supervising the new with the old: learning sfm from sfm.
\newblock In {\em Proceedings of the European Conference on Computer Vision
  (ECCV)}, pages 698--713, 2018.

\bibitem{kornblith2019better}
Simon Kornblith, Jonathon Shlens, and Quoc~V Le.
\newblock Do better imagenet models transfer better?
\newblock In {\em Proceedings of the IEEE/CVF Conference on Computer Vision and
  Pattern Recognition (CVPR)}, pages 2661--2671, 2019.

\bibitem{kosaka1992fast}
Akio Kosaka and Avinash~C Kak.
\newblock Fast vision-guided mobile robot navigation using model-based
  reasoning and prediction of uncertainties.
\newblock {\em CVGIP: Image understanding}, 56(3):271--329, 1992.

\bibitem{krizhevsky2012imagenet}
Alex Krizhevsky, Ilya Sutskever, and Geoffrey~E Hinton.
\newblock Imagenet classification with deep convolutional neural networks.
\newblock {\em Advances in Neural Information Processing Systems (NeurIPS)},
  25:1097--1105, 2012.

\bibitem{kwon2020uncertainty}
Yongchan Kwon, Joong-Ho Won, Beom~Joon Kim, and Myunghee~Cho Paik.
\newblock Uncertainty quantification using bayesian neural networks in
  classification: Application to biomedical image segmentation.
\newblock {\em Computational Statistics \& Data Analysis}, 142:1--17, 2020.

\bibitem{ladicky2014pulling}
Lubor Ladicky, Jianbo Shi, and Marc Pollefeys.
\newblock Pulling things out of perspective.
\newblock In {\em Proceedings of the IEEE Conference on Computer Vision and
  Pattern Recognition (CVPR)}, pages 89--96, 2014.

\bibitem{lakshminarayanan2017simple}
Balaji Lakshminarayanan, Alexander Pritzel, and Charles Blundell.
\newblock Simple and scalable predictive uncertainty estimation using deep
  ensembles.
\newblock In {\em Proceedings of the 31st International Conference on Neural
  Information Processing Systems (NeurIPS)}, page 6405–6416. Curran
  Associates Inc., 2017.

\bibitem{li2010towards}
Congcong Li, Adarsh Kowdle, Ashutosh Saxena, and Tsuhan Chen.
\newblock Towards holistic scene understanding: Feedback enabled cascaded
  classification models.
\newblock In {\em Advances in Neural Information Processing Systems (NeurIPS)},
  pages 1351--1359, 2010.

\bibitem{li2020unsupervised}
Hanhan Li, Ariel Gordon, Hang Zhao, Vincent Casser, and Anelia Angelova.
\newblock Unsupervised monocular depth learning in dynamic scenes.
\newblock {\em arXiv preprint arXiv:2010.16404}, 2020.

\bibitem{li2019learning}
Zhengqi Li, Tali Dekel, Forrester Cole, Richard Tucker, Noah Snavely, Ce Liu,
  and William~T Freeman.
\newblock Learning the depths of moving people by watching frozen people.
\newblock In {\em Proceedings of the IEEE/CVF Conference on Computer Vision and
  Pattern Recognition (CVPR)}, pages 4521--4530, 2019.

\bibitem{luo2020every}
Chenxu Luo, Zhenheng Yang, Peng Wang, Yang Wang, Wei Xu, Ram Nevatia, and Alan
  Yuille.
\newblock Every pixel counts++: Joint learning of geometry and motion with 3d
  holistic understanding.
\newblock {\em IEEE Transactions on Pattern Analysis and Machine Intelligence},
  42(10):2624--2641, 2020.

\bibitem{ma2019self}
Fangchang Ma, Guilherme~Venturelli Cavalheiro, and Sertac Karaman.
\newblock Self-supervised sparse-to-dense: Self-supervised depth completion
  from lidar and monocular camera.
\newblock In {\em 2019 International Conference on Robotics and Automation
  (ICRA)}, pages 3288--3295. IEEE, 2019.

\bibitem{ma2018shufflenet}
Ningning Ma, Xiangyu Zhang, Hai-Tao Zheng, and Jian Sun.
\newblock Shufflenet v2: Practical guidelines for efficient cnn architecture
  design.
\newblock In {\em Proceedings of the European conference on computer vision
  (ECCV)}, pages 116--131, 2018.

\bibitem{mahjourian2018unsupervised}
Reza Mahjourian, Martin Wicke, and Anelia Angelova.
\newblock Unsupervised learning of depth and ego-motion from monocular video
  using 3d geometric constraints.
\newblock In {\em Proceedings of the IEEE Conference on Computer Vision and
  Pattern Recognition (CVPR)}, pages 5667--5675, 2018.

\bibitem{meng2019signet}
Yue Meng, Yongxi Lu, Aman Raj, Samuel Sunarjo, Rui Guo, Tara Javidi, Gaurav
  Bansal, and Dinesh Bharadia.
\newblock Signet: Semantic instance aided unsupervised 3d geometry perception.
\newblock In {\em Proceedings of the IEEE/CVF Conference on Computer Vision and
  Pattern Recognition (CVPR)}, pages 9810--9820, 2019.

\bibitem{mnih2014neural}
Andriy Mnih and Karol Gregor.
\newblock Neural variational inference and learning in belief networks.
\newblock In {\em International Conference on Machine Learning (ICML)}, pages
  1791--1799. PMLR, 2014.

\bibitem{mukhoti2018evaluating}
Jishnu Mukhoti and Yarin Gal.
\newblock Evaluating bayesian deep learning methods for semantic segmentation.
\newblock {\em arXiv preprint arXiv:1811.12709}, 2018.

\bibitem{nair2010rectified}
Vinod Nair and Geoffrey~E Hinton.
\newblock Rectified linear units improve restricted boltzmann machines.
\newblock In {\em Proceedings of the 27th International Conference on
  International Conference on Machine Learning (ICML)}, pages 807--814, 2010.

\bibitem{poggi2020uncertainty}
Matteo Poggi, Filippo Aleotti, Fabio Tosi, and Stefano Mattoccia.
\newblock On the uncertainty of self-supervised monocular depth estimation.
\newblock In {\em Proceedings of the IEEE/CVF Conference on Computer Vision and
  Pattern Recognition (CVPR)}, pages 3227--3237, 2020.

\bibitem{ranjan2019competitive}
Anurag Ranjan, Varun Jampani, Lukas Balles, Kihwan Kim, Deqing Sun, Jonas
  Wulff, and Michael~J Black.
\newblock Competitive collaboration: Joint unsupervised learning of depth,
  camera motion, optical flow and motion segmentation.
\newblock In {\em Proceedings of the IEEE Conference on Computer Vision and
  Pattern Recognition (CVPR)}, pages 12240--12249, 2019.

\bibitem{ren2017unsupervised}
Zhe Ren, Junchi Yan, Bingbing Ni, Bin Liu, Xiaokang Yang, and Hongyuan Zha.
\newblock Unsupervised deep learning for optical flow estimation.
\newblock {\em Proceedings of the AAAI Conference on Artificial Intelligence},
  31(1):1495--1501, 2017.

\bibitem{ritter2018scalable}
Hippolyt Ritter, Aleksandar Botev, and David Barber.
\newblock A scalable laplace approximation for neural networks.
\newblock In {\em 6th International Conference on Learning Representations,
  ICLR 2018-Conference Track Proceedings}, volume~6. International Conference
  on Representation Learning, 2018.

\bibitem{sandler2018mobilenetv2}
Mark Sandler, Andrew Howard, Menglong Zhu, Andrey Zhmoginov, and Liang-Chieh
  Chen.
\newblock Mobilenetv2: Inverted residuals and linear bottlenecks.
\newblock In {\em Proceedings of the IEEE Conference on Computer Vision and
  Pattern Recognition (CVPR)}, pages 4510--4520, 2018.

\bibitem{saxena2008make3d}
Ashutosh Saxena, Min Sun, and Andrew~Y Ng.
\newblock Make3d: Learning 3d scene structure from a single still image.
\newblock {\em IEEE Transactions on Pattern Analysis and Machine Intelligence},
  31(5):824--840, 2008.

\bibitem{szegedy2015going}
Christian Szegedy, Wei Liu, Yangqing Jia, Pierre Sermanet, Scott Reed, Dragomir
  Anguelov, Dumitru Erhan, Vincent Vanhoucke, and Andrew Rabinovich.
\newblock Going deeper with convolutions.
\newblock In {\em Proceedings of the IEEE Conference on Computer Vision and
  Pattern Recognition (CVPR)}, pages 1--9, 2015.

\bibitem{szeliski1990bayesian}
Richard Szeliski.
\newblock Bayesian modeling of uncertainty in low-level vision.
\newblock {\em International Journal of Computer Vision}, 5(3):271--301, 1990.

\bibitem{tan2019mnasnet}
Mingxing Tan, Bo Chen, Ruoming Pang, Vijay Vasudevan, Mark Sandler, Andrew
  Howard, and Quoc~V Le.
\newblock Mnasnet: Platform-aware neural architecture search for mobile.
\newblock In {\em Proceedings of the IEEE/CVF Conference on Computer Vision and
  Pattern Recognition (CVPR)}, pages 2820--2828, 2019.

\bibitem{tan2019efficientnet}
Mingxing Tan and Quoc Le.
\newblock Efficientnet: Rethinking model scaling for convolutional neural
  networks.
\newblock In {\em International Conference on Machine Learning (ICML)}, pages
  6105--6114. PMLR, 2019.

\bibitem{wan2020fbnetv2}
Alvin Wan, Xiaoliang Dai, Peizhao Zhang, Zijian He, Yuandong Tian, Saining Xie,
  Bichen Wu, Matthew Yu, Tao Xu, Kan Chen, et~al.
\newblock Fbnetv2: Differentiable neural architecture search for spatial and
  channel dimensions.
\newblock In {\em Proceedings of the IEEE/CVF Conference on Computer Vision and
  Pattern Recognition (CVPR)}, pages 12965--12974, 2020.

\bibitem{wang2017stereo}
Rui Wang, Martin Schworer, and Daniel Cremers.
\newblock Stereo dso: Large-scale direct sparse visual odometry with stereo
  cameras.
\newblock In {\em Proceedings of the IEEE International Conference on Computer
  Vision (ICCV)}, pages 3903--3911, 2017.

\bibitem{yang2020d3vo}
Nan Yang, Lukas~von Stumberg, Rui Wang, and Daniel Cremers.
\newblock D3vo: Deep depth, deep pose and deep uncertainty for monocular visual
  odometry.
\newblock In {\em Proceedings of the IEEE/CVF Conference on Computer Vision and
  Pattern Recognition (CVPR)}, pages 1281--1292, 2020.

\bibitem{yang2018netadapt}
Tien-Ju Yang, Andrew Howard, Bo Chen, Xiao Zhang, Alec Go, Mark Sandler,
  Vivienne Sze, and Hartwig Adam.
\newblock Netadapt: Platform-aware neural network adaptation for mobile
  applications.
\newblock In {\em Proceedings of the European Conference on Computer Vision
  (ECCV)}, pages 285--300, 2018.

\bibitem{yin2018geonet}
Zhichao Yin and Jianping Shi.
\newblock Geonet: Unsupervised learning of dense depth, optical flow and camera
  pose.
\newblock In {\em Proceedings of the IEEE Conference on Computer Vision and
  Pattern Recognition (CVPR)}, pages 1983--1992, 2018.

\bibitem{ying2019bench}
Chris Ying, Aaron Klein, Eric Christiansen, Esteban Real, Kevin Murphy, and
  Frank Hutter.
\newblock Nas-bench-101: Towards reproducible neural architecture search.
\newblock In {\em International Conference on Machine Learning (ICML)}, pages
  7105--7114. PMLR, 2019.

\bibitem{zhou2017unsupervised}
Tinghui Zhou, Matthew Brown, Noah Snavely, and David~G Lowe.
\newblock Unsupervised learning of depth and ego-motion from video.
\newblock In {\em Proceedings of the IEEE Conference on Computer Vision and
  Pattern Recognition (CVPR)}, pages 1851--1858, 2017.

\bibitem{zhu2019deformable}
Xizhou Zhu, Han Hu, Stephen Lin, and Jifeng Dai.
\newblock Deformable convnets v2: More deformable, better results.
\newblock In {\em Proceedings of the IEEE/CVF Conference on Computer Vision and
  Pattern Recognition (CVPR)}, pages 9308--9316, 2019.

\bibitem{zou2018df}
Yuliang Zou, Zelun Luo, and Jia-Bin Huang.
\newblock Df-net: Unsupervised joint learning of depth and flow using
  cross-task consistency.
\newblock In {\em Proceedings of the European Conference on Computer Vision
  (ECCV)}, pages 36--53, 2018.

\end{thebibliography}
}

\clearpage
\appendix
\part*{Supplementary Material}
\section{Qualitative Results}

\subsection{Motion Maps}
\begin{figure}[ht!]
\begin{center}
   \includegraphics[width=1\linewidth]{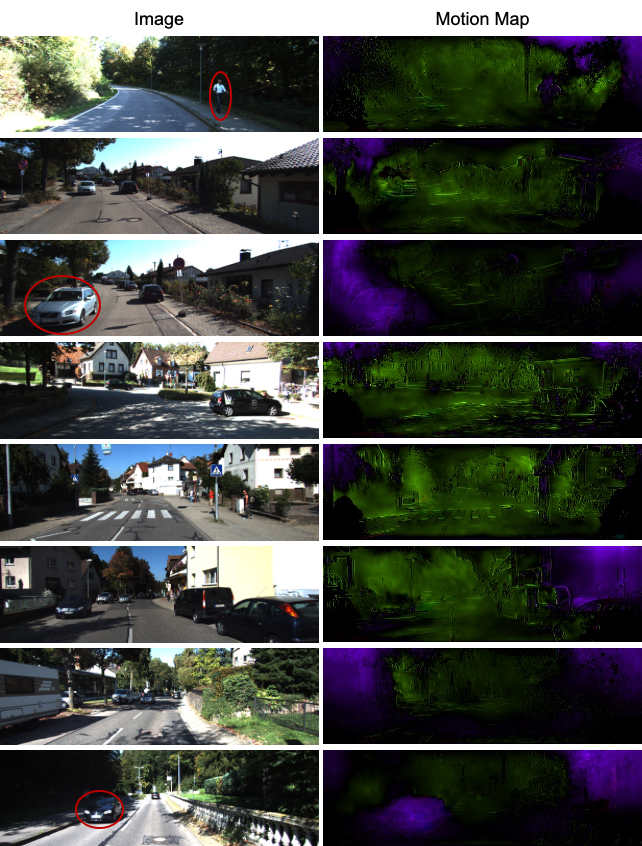}
\end{center}
   \caption{Visualization of the motion maps learned. The learned motion map compensates for large dynamic objects that are close to the camera. The red circles mark the dynamic objects in the scenes.}
\label{fig:qualitative_results_motion}
\end{figure}

Fig. \ref{fig:qualitative_results_motion} displays the motion maps estimated for the KITTI odometry dataset. The motion map compensates for large dynamic objects that are close to the camera. In addition, the motion map not only handles dynamic objects but also deals with the areas tricky for photometric reconstruction. For example, the motion map estimates motions for the areas with leaves or the part of buildings where the sun is shining down. These results imply that the motion map formulation aids the depth estimation network in learning to compensate for the violation of the implicit assumptions of the photometric consistency loss: the static scene assumption and the Lambertian surface assumption.

\subsection{Depth Maps}
\begin{figure*}[t]
\begin{center}
   \includegraphics[width=1\linewidth]{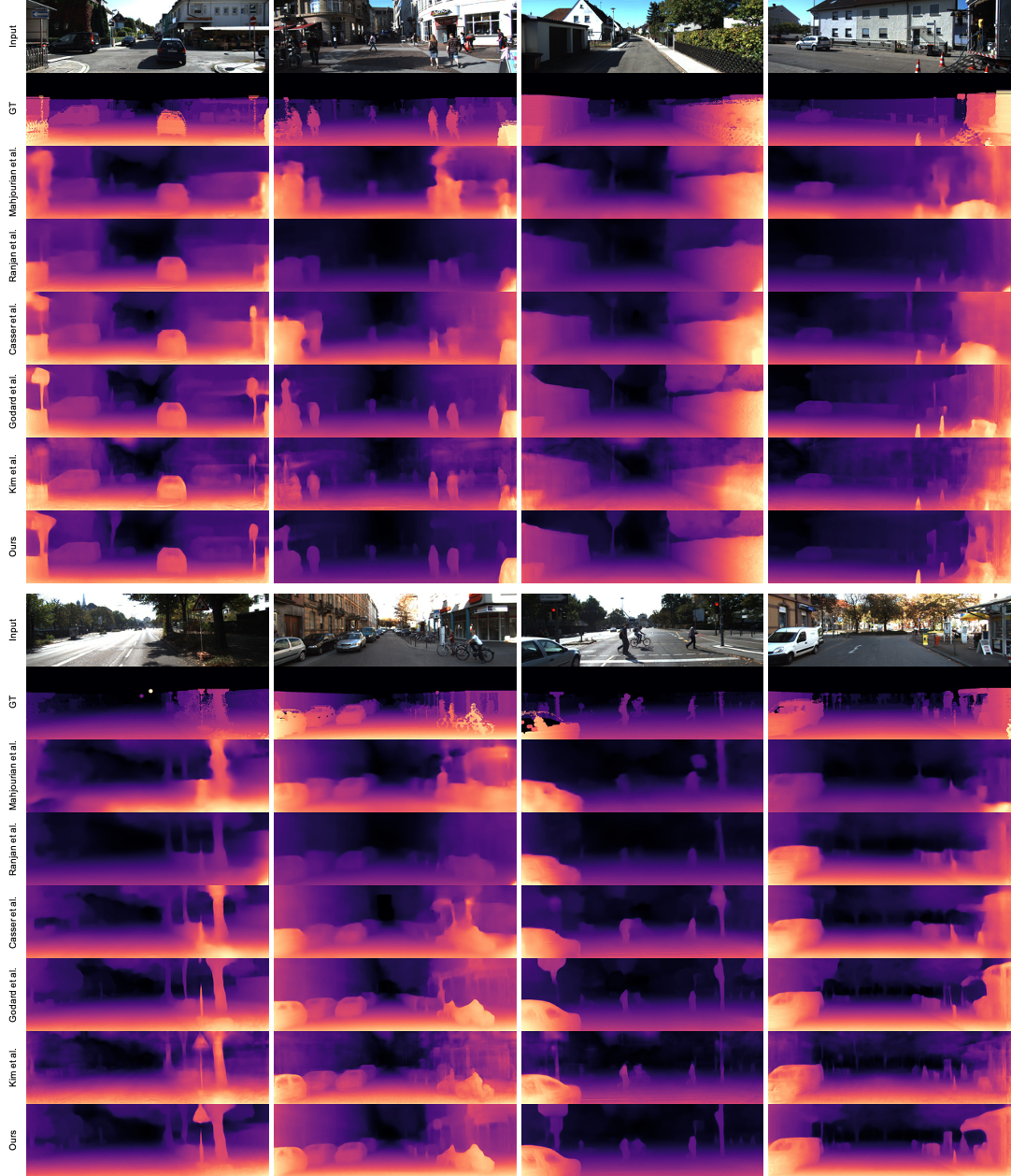}
\end{center}
   \caption{The qualitative comparison of depths maps. The depth maps of the proposed method clearly delineate the outlines of objects. Compared to the interpolated ground-truth depth maps, the proposed method provides continuous depth maps without holes. Moreover, the depth maps of the proposed method could handle the dynamic objects such as humans and vehicles.}
\label{fig:qualitative_results}
\end{figure*}

Fig. \ref{fig:qualitative_results} compares the depth maps estimated by our method to those of conventional methods. In most cases, the depth maps of the proposed method display more fine-grained structures of objects and scenes. Moreover, the depth maps of our method delineate the outline of dynamic objects such as humans and vehicles in a concrete manner. Compared to the interpolated ground-truth depth maps, the depth maps from the proposed method show continuous depth values without holes.

\section{Comparative Study}
Table \ref{tb:result_quantitative_depth} quantitatively compares the depth estimation results of the proposed method to conventional methods. Recent methods have employed either the DispNet architecture or the ResNet-18 architecture, while two of them employed ResNet-50 and ResNet-101. For a fair comparison, we indicate the architecture each method has employed. The proposed method displays superior or on-par performance compared to the state-of-the-art models according to the performance metrics. Here, we emphasize once again that the purpose of our study is to investigate (1) the inter-dependency between recent learning methods for self-supervised depth and motion learning and (2) the effect of architectural factors on performance. The performance gain followed as a result of our study. Besides, the need to search for the optimal architecture for self-supervised depth and motion learning lingers though we have investigated the architectural impact on the performance.

\begin{table*}[t]
\caption{Quantitative comparison results. The best results are in bold and the second best results are underlined. The proposed method displays superior or on-par performance compared to the state-of-the-art models.}
\label{tb:result_quantitative_depth}
\vskip 0.05in
\begin{center}
\begin{tabular}{llccccccc}
\toprule
\multicolumn{1}{c}{\textbf{Method}} & \multicolumn{1}{c}{\textbf{Architecture}} & \textbf{ARD} & \textbf{SRD} & \textbf{RMSE} & \textbf{RMSE log} & $\delta \!<\! 1.25$
& $\delta \!<\! 1.25^2$
& $\delta \!<\! 1.25^3$\\
\midrule
Mahjourian \emph{et al.} \cite{mahjourian2018unsupervised} & DispNet \cite{zhou2017unsupervised} & 0.163 & 1.240 & 6.220 & 0.250 & 0.762 & 0.916 & 0.968\\
Ranjan \emph{et al.} \cite{ranjan2019competitive} & DispNet \cite{zhou2017unsupervised} & 0.148 & 1.149 & 5.464 & 0.226 & 0.815 & 0.935 & 0.973\\
Casser \emph{et al.} \cite{casser2019depth} & ResNet-18 & 0.141 & 1.026 & 5.291 & 0.215 & 0.816 & 0.945 & 0.979\\
Godard \emph{et al.} \cite{godard2019digging} & ResNet-18 & 0.115 & 0.903 & 4.863 & 0.193 & 0.877 & 0.959 & 0.981\\
Kim \emph{et al.} \cite{kim2020simvodis} & ResNet-50 & 0.123 & \underline{0.797} & 4.727 & 0.193 & 0.854 & 0.960 & \textbf{0.984}\\
Johnston \emph{et al.} \cite{johnston2020self} & ResNet-18 & 0.111 & 0.941 & 4.817 & 0.189 & \underline{0.885} & \underline{0.961} & 0.981\\
Johnston \emph{et al.} \cite{johnston2020self} & ResNet-101 & \textbf{0.106} & 0.861 & \underline{4.699} & \textbf{0.185} & \textbf{0.889} & \textbf{0.962} & \underline{0.982}\\
\midrule
Ours & ResNet-18 & 0.114 & 0.825 & 4.706 & 0.191 & 0.877 & 0.960 & 0.982\\
Ours & DeResNet-50 & \underline{0.108} & \textbf{0.737} & \textbf{4.562} & \underline{0.187} & 0.883 & \underline{0.961} & \underline{0.982}\\
\bottomrule
\end{tabular}
\end{center}
\end{table*}

\end{document}